\documentclass{article}

\usepackage[nonatbib,preprint]{neurips_2023}

\usepackage[utf8]{inputenc} 
\usepackage[T1]{fontenc}    
\usepackage[hidelinks]{hyperref}       
\usepackage{url}            
\usepackage{booktabs}       
\usepackage{amsfonts}       
\usepackage{nicefrac}       
\usepackage{microtype}      
\usepackage{xcolor}         

\usepackage[backend=bibtex,autocite=superscript]{biblatex}
\bibliography{bibli}
\usepackage{graphicx}
\usepackage{amsmath}
\usepackage{amssymb}

\newtheorem{theorem}{Theorem}

\title{Understanding polysemanticity in neural networks through coding theory}

\author{%
  Simon C. Marshall\thanks{s.c.marshall@liacs.leidenuniv.nl} \\
  LIACS, Leiden University \\
  \And
  Jan H. Kirchner \\
  OpenAI\thanks{Work done outside of OpenAI.} \\
}

\begin{document}

\maketitle

\begin{abstract}
Despite substantial efforts, neural network interpretability remains an elusive goal, with previous research failing to provide succinct explanations of most single neurons' impact on the network output. This limitation is due to the polysemantic nature of most neurons, whereby a given neuron is involved in multiple unrelated network states, complicating the interpretation of that neuron. In this paper, we apply tools developed in neuroscience and information theory to propose both a novel practical approach to network interpretability and theoretical insights into polysemanticity and the density of codes. We infer levels of redundancy in the network's code by inspecting the eigenspectrum of the activation's covariance matrix. Furthermore, we show how random projections can reveal whether a network exhibits a smooth or non-differentiable code and hence how interpretable the code is. This same framework explains the advantages of polysemantic neurons to learning performance and explains trends found in recent results by Elhage et al.~(2022).
Our approach advances the pursuit of interpretability in neural networks, providing insights into their underlying structure and suggesting new avenues for circuit-level interpretability.
\end{abstract}

\section{Introduction}
As the capabilities of deep neural networks increase, the need for interpretability to understand failure modes, fix bugs, and establish accountability becomes increasingly apparent\autocite{doshi2017towards,lipton2018mythos,raukur2022toward}. Given an explanation of the internal structure of a deep neural network, we might be able to infer the logic that leads to a given output and correct this logic to conform with our preferences for the model's behaviour. The methods for achieving interpretability overlap with those of other areas of deep learning research. For example, more interpretable networks tend to be more robust to adversarial attacks\autocite{ross2018improving,finlay2019scaleable}, while more robust networks tend to be more interpretable\autocite{engstrom2019adversarial,augustin2020adversarial}. Modularity and continual learning are also related to interpretability, as methods that promote specialization among weights or neurons can make the network easier to understand\autocite{kirkpatrick2017overcoming,li2017learning,rusu2016progressive}.

Despite the rapid development of interpretability techniques, key challenges remain. Techniques need to be sufficiently robust to scale to large models\autocite{patrick2022capsule,raukur2022toward}, withstand rigorous evaluation\autocite{hubinger2021automating,holmberg2022towards}, and work reliably across domains\autocite{ding2021grounding}. One particular challenge for interpretability is the phenomenon of polysemanticity, whereby neurons are activated by multiple unrelated features\autocite{fong2018net2vec,mu2020compositional}. Polysemanticity poses a challenge for interpretability based on individual neurons\autocite{bills2023language}, which will necessarily be incorrect or incomplete. One explanation for why polysemanticity is near ubiquitous is the \textit{superposition hypothesis} which posits that neural networks overload individual neurons with multiple roles in order to leverage the maximum available network capacity\autocite{olah2020zoom,elhage2022toy}. A deep understanding of these phenomena will be important for advancing interpretability research.

In parallel with developments in interpretability research, information theory has emerged as a promising approach for analyzing neural networks. A landmark study in this field proposed to analyze deep neural networks by studying the plane formed by the amount of mutual information between hidden activations and the network's input and output variables\autocite{tishby2015deep}. Applying this perspective to network activations during training provides deep insight into the trade-offs and learning dynamics that shape emerging representations\autocite{shwartz2017opening}. Intriguingly, this work suggests that many aspects of deep neural networks can be explained solely by optimality considerations and without reference to the details of training \autocite{tishby2015deep,huang2019information}, suggesting a possible top-down approach to studying network interpretability.

Building on these previous works, in this paper, we propose a novel approach to interpretability by analyzing the eigenspectrum of neural networks from a coding perspective. We leverage the decay rate of the eigenspectrum of the activation covariance matrix to infer the amount of redundancy in the neural network's code. This approach offers insights into the code's high-level properties, including its channel capacity and information flow. We demonstrate how random projections can be employed to determine whether a network exhibits a smooth or non-differentiable phase transition, further enhancing our understanding of the network's structure and behaviour. Our approach has the potential to inform the design of more interpretable and robust neural networks, ultimately leading to more transparent and reliable deep learning models.

\section{Results}
\begin{figure}
    \centering
    \includegraphics{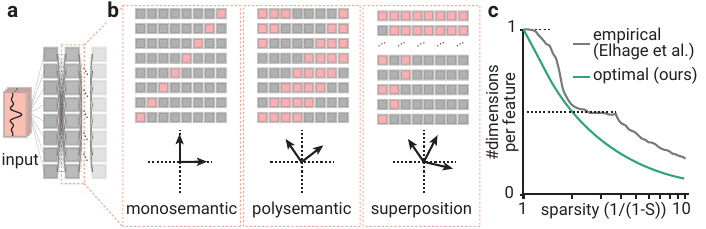}
    \caption{\textbf{Network capacity of deep artificial neural networks.} \textbf{a.} Schematic of deep artificial neural network (ANN) receiving a set of input signals (left; orange) and processing the signal through consecutive layers of neurons (right; shades of gray) connected in an all-to-all fashion. \textbf{b.} Schematic of different possible coding schemes within a layer of the ANN. \textit{Monosemantic coding} (left): one-to-one identification between inputs and neuron activations (highlighted in orange) without overlap. \textit{Polysemantic coding} (middle): One-to-one identification between inputs and neuron activations \textit{with} overlap. \textit{Superposition coding} (right): One-to-many identification between inputs and neuron activations. Note that the third scenario is a sub-type of the second scenario. \textbf{c.} Number of dimensions needed in a neural network's hidden layer per input feature as a function of varying feature sparsity(gray) compared to our theoretical prediction of optimal encoding (green). Data extracted from ref.~\cite{elhage2022toy}. Dashed horizontal lines indicate "sticky" regions\autocite{elhage2022toy} where the network code deviates from optimal.}
    \label{fig:schematic}
\end{figure}

\textbf{Polysemanticity and superposition as the default for efficient encoding of information.} When presented with an appropriately formatted input, the hidden layers of a deep neural network activate based on the characteristics of the input and based on the connectivity of neurons within that network (Fig.~\ref{fig:schematic}a). Individual neurons in the network will activate very selectively to a few inputs (monosemantic coding) or will activate broadly to multiple inputs (polysemantic/superposition coding) depending on the learned code (Fig.~\ref{fig:schematic}b). While these phenomena traditionally are studied at the single-neuron level\autocite{elhage2021mathematical}, we regard them as properties of the network code. Viewed from this perspective, it is immediately apparent that a network that needs to encode a number of features larger than the number of neurons per layer will benefit from utilizing a superposition code (Fig.~\ref{fig:schematic}b).

To investigate whether neural networks can indeed learn to leverage an approximately optimal encoding of its input, we reanalyze a result from a recent study on toy models of superposition\autocite{elhage2022toy}. In this setting, a deep neural network learns to encode and decode inputs, $x$, with varying sparsity, $S$. By computing the information capacity of a neural network and comparing it to the information content of a sparse input, we can calculate the compression rate, $R$, that could be achieved by an optimal code,

\begin{align}
    R = 1-S + \frac{1}{H}(-S \log(S) - (1-S)\log(1-S)),
\end{align}


where $H = \langle \log p_x \rangle$ is the information content (entropy) of a single random input, $x$ (see Supplementary Note for detailed derivation). For sufficiently large $H$, e.g.~16 or 32 bits, $R$ can be approximated as $R=1-S$. Intriguingly, this analytic result provides a good fit to the simulation data (Fig.\ref{fig:schematic}c), with the deviations from the optimal being due to ``sticky" encoding regimes where the toy model learns a highly robust but suboptimal encoding\autocite{elhage2022toy}.

Might an optimal code be a more interpretable code? The hardness of average case decoding of neural network codes is unknown, but results for specific tasks indicate the problem might be difficult\autocite{goldwasser2022planting}. Indeed, for some tasks, we demonstrate that the optimal code resists polynomial time decoders almost always (see Supplementary Note). Given that optimal polysemantic codes reuse the same neuron for many concepts and deciding properties of highly convoluted functions is generally hard, we might expect networks that learn a nearly optimal code to be very difficult to interpret.

\begin{figure}
    \centering
    \includegraphics{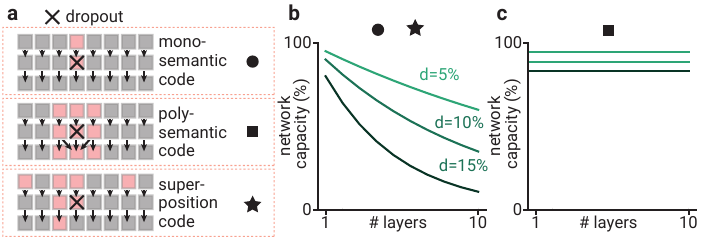}
    \caption{\textbf{Redundancy in channel code enables noise robustness.} \textbf{a.} Illustration of the effect of dropout on the three types of code introduced in Figure \ref{fig:schematic}. While dropout corrupts information in the monosemantic (top, circle) and superposition (bottom, star) codes, a polysemantic code can recover from dropout. \textbf{b., c.} Network capacity as a function of network for different levels of dropout $\alpha$ for monosemantic (\textbf{b}, circle), superposition (\textbf{b}, star) and polysemantic code (\textbf{c}, square).}
    \label{fig:error}
\end{figure}

\textbf{Robust encoding implies reduced channel capacity.} While the code learned by networks might be very difficult to interpret in general due to its high information content, in practice, neural networks do not solely maximize information density. In the face of noise and other perturbations, neural networks must be robust by introducing redundancies and other error-correcting steps into the code. To illustrate the effect of noise, we consider dropout\autocite{srivastava2014dropout}, a common form of regularisation in neural networks (Fig.~\ref{fig:error}a). Dropout is itself a noise source: as information passes through layers, a fraction, $D\in[0,1]$, of the total number of neurons is erased. A naïve dense code (such as that from the previous subsection) with no redundancies would immediately lose a fraction $D$ of information encoded across neurons in the first dropout layer and then a fraction $D$ again in the subsequent dropout layers, giving an effective information capacity of $\Tilde{C}_{\mathit{dropout}}\sim C (1-D)^N$, after $N$ dropout layers, where $C$ is the capacity of the noiseless channel (Fig.~\ref{fig:error}b). 

However, by utilizing a redundant encoding, it is possible to introduce an error-correcting component into the code, which is resistant to some noise sources. When error-correction is present, the neural network can reliably transmit the entire channel capacity of a layer with dropout through every layer of the network (Fig.~\ref{fig:error}c, see Supplementary Note for detailed derivation),

\begin{align}
    C_{\mathit{dropout}}\sim C (1-D).
\end{align}

We verified that a neural network trained as an autoencoder could indeed learn error-correction by measuring the mutual information between the input and the hidden layers\autocite{belghazi2018mine} and observing that after training, the network does not experience exponential decay of information (see Supplementary Note for details).

Thus, when a neural network is exposed to moderate noise, we expect the code to become more redundant, distributing the representation of inputs across multiple activation patterns. While this distribution of representations across neurons discourages a monosemantic code (Fig.~\ref{fig:error}a), it is unclear whether it might facilitate the holistic interpretation of activations across a given layer.

\begin{figure}
    \centering
    \includegraphics{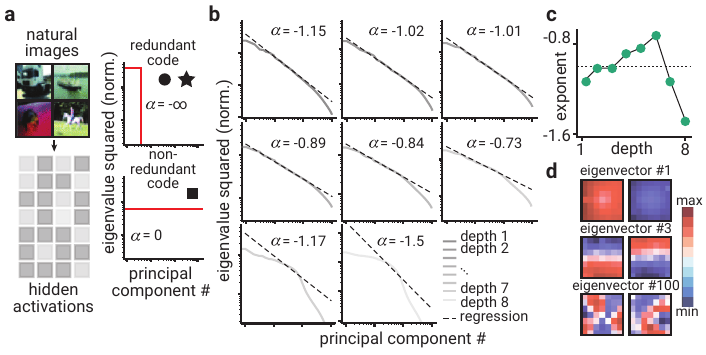}
    \caption{\textbf{The code shapes hidden activation eigenspectrum.} \textbf{a.} Left: Schematic illustrating input to (natural images from the CIFAR dataset) and hidden activations of an Autoencoder-ResNet (see Methods for details). Right: Idealized eigenspectrum for hidden activations implementing a maximally redundant (top) or maximally non-redundant (bottom) code. \textbf{b.} Log-log plot of eigenvalues as a function of principal components for hidden activations from different network depths (shades of grey and legend in bottom right). Dashed line indicates the power law fit obtained from Huber regression (see Methods). Large deviations from the power law distribution exist for later layers, hence power law fits represent approximations. Note that the later eigenspectra resemble the predicted spectrum of a linear code (compare \textbf{a}), where the code words are simple linear combinations. \textbf{c.} Estimated power law exponent $\alpha$ as a function of network depth. Dashed line marks $\alpha=-1$. \textbf{d.} Sample of different eigenvectors (rows) displayed as filters of the ResNet.}
    \label{fig:spectrum}
\end{figure}

\textbf{Codes produce eigenspectra with varying speeds of decay.} We next turned to probe the high-level properties of a neural network's code. 
A natural approach that was recently proposed in the study of coding properties of biological systems\autocite{stringer2019high,kong2022increasing} is to investigate the eigenspectrum of the activation covariance matrix. 
The shape of the eigenspectrum can reflect the properties of the network's code.
In particular, a largely redundant code has an eigenspectrum with a steep decay since only a small subset of possible configurations is actually relevant to the network's code (Fig.~\ref{fig:spectrum}a). 
In contrast, a maximally non-redundant code will exhibit a more gradual decay in the eigenspectrum since it exploits the entire space of possible configurations (Fig.~\ref{fig:spectrum}a).
In practice, eigenspectra tend not to match either extreme and instead decay according to a power law with parameter $\alpha$. Applying this approach to deep neural networks, we find the same power law decay but with varying values of $\alpha$.
By using this encoding, the effective channel capacity becomes approximately.

\begin{align}
    C_\alpha \approx C- |\alpha| m \log(m/e),
\end{align}

where $m$ is the width of the given layer and $C$ is the channel capacity using the optimal dense encoding (see Supplementary Note for detailed derivation).

When visualizing the eigenspectrum of the hidden activations of a trained Autoencoder-ResNet trained on the ImageNet dataset without dropout (Fig.~\ref{fig:spectrum}a), we find that the decay of the eigenspectrum changes non-monotonically over the depth of the network (Fig.~\ref{fig:spectrum}b). The eigenspectrum initially decays with a power law exponent just below $-1$, which is consistent with the observation that the spectrum of natural images decays with that same exponent\autocite{stringer2019high}. The eigenspectrum decay then slows down, with the power law exponent increasing above $-1$ for the intermediate layers of the network, before eventually decaying rapidly with a power law exponent $\leq -1$ for the final layers  (Fig.~\ref{fig:spectrum}c). Consistent with results from neuroscience\autocite{stringer2019high}, the corresponding eigenfilters vary from coarse global filters for the first eigenvectors to intricate filters sensitive to local variations for later eigenvectors. We note that the recovered eigenvectors resemble previously described Gabor receptive fields\autocite{kruger2012deep} as well as curve detectors\autocite{cammarata2020thread} (Fig.~\ref{fig:spectrum}d).

In summary, the eigenspectrum of a trained Autoencoder-ResNet reveals that the trained network learns a code that is neither highly redundant nor one that maximizes capacity. Instead, we find that the neural code exhibits some redundancy, particularly towards early and late layers, while employing less redundant codes in intermediate layers.

\begin{figure}
    \centering
    \includegraphics{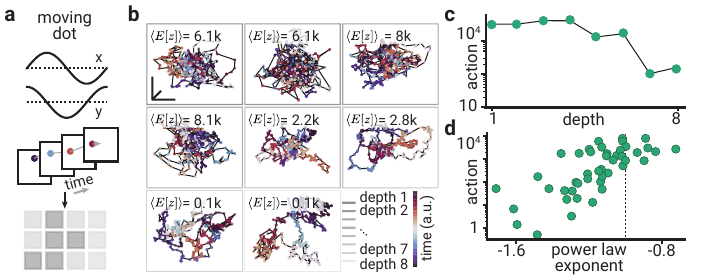}
    \caption{\textbf{Random projections of hidden activations exhibit varying amount of smoothness.} \textbf{a.} Simple moving dot stimulus generated from the $\sin$ and $\cos$ function (top), converted into a stack of frames (middle), and passed into the network to yield hidden activations (bottom). \textbf{b.} Sample of random projections of hidden activations produced from the moving dot stimulus. Color of box indicates layer of the network, and color of the random projection indicates time. Inset text shows average activation. \textbf{c.} Average action of random projections as a function of network depth. Average computed over 1000 random projections (see Methods). \textbf{d.} Average action of random projections as a function of the estimated power law exponent $\alpha$.}
    \label{fig:projection}
\end{figure}

\textbf{Random projections reflect smoothness of code.} What can we say about the neural network code beyond general statements regarding its redundancy? We hypothesized that an input that is encoded with more redundancy might be easier to decode with linear probes\autocite{alain2016understanding}. 
To this end, we projected the $m$ hidden activations produced by a moving dot stimulus with $n$ frames (Fig.~\ref{fig:projection}a), $A\in\mathbb{R}^{n\times m}$, onto a randomly chosen basis of vectors with normalized variance, $r\in\mathbb{R}^{m \times 3}, r_i\sim\mathcal{N}(0,\frac{1}{m}I)$. These random projections, $z = Ar$, have the useful property that their average action, $\langle E[z] \rangle = \frac{1}{2}\int ||\dot z(t)||^2 \mathrm{d}t$, is proportional to the decay factor $\alpha$,

\begin{align}
    \langle E[z] \rangle = \sum_{k=1}^{m} \lambda_k \sim \frac{1}{2} \zeta({-\alpha}),\label{zetaequation}
\end{align}

where $\lambda_i$ are the eigenvalues of the hidden activations and $\zeta$ is the Riemann zeta function restricted to the real numbers, $\zeta (s) = \sum_{k=1}^\infty \frac{1}{k^s}$.

When displaying the random projections as curves, we indeed observed that while early layers produce highly discontinuous projections, later layers of the network recover some of the smooth structure of the input (Fig.~\ref{fig:projection}b). In particular, we noticed that the transition between discontinuous and locally smooth occurs as the decay factor $\alpha$ falls below $-1$  (Fig.~\ref{fig:projection}c). From eq.~\ref{zetaequation}, we see that in this regime, the expected action of the curve is finite, while in the regime $\alpha > -1$, the action of the curve grows without bound.

The slow decay of eigenspectra in artificial neural networks stands in stark contrast to the eigenspectra in the visual cortex of the mouse\autocite{stringer2019high} and the macaque\autocite{kong2022increasing}. In this biological network, the power law exponent $\alpha$ is consistently below $-1$, constraining the network to use a continuous and differential code of the input. Instead, the artificial neural network resembles the much smaller zebrafish\autocite{wang2023scale} with a slowly decaying eigenspectrum with $\alpha \approx - 0.7$.

Taken together, we found that the speed of decay of the eigenspectrum determines the statistical properties of random projections of stimuli. A fast decay ($\alpha \leq -1$) is common in larger mammals\autocite{stringer2019high,kong2022increasing} and results in continuous and differentiable projections. A slow decay ($\alpha > -1$) is more prominent in small artificial neural networks and in the zebrafish\autocite{wang2023scale} and implies discontinuous random projections.

\begin{figure}
    \centering
    \includegraphics{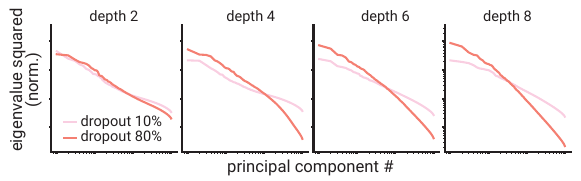}
    \caption{\textbf{Higher noise induces more robust codes}. Log-log plot of eigenvalues as a function of principal components for hidden activations for two different networks with different dropout rates. The increase in non-linearity indicates a learnt code which is more robust to noise (which dropout introduces).}
    \label{fig:Noisy is non-linear}
\end{figure}

\textbf{Deviations from power law distribution}
In the final layers of figure \ref{fig:spectrum} we can see that the eigenspectrum in some cases systematically deviates from the power law. These deviations stem from transitions from the `biological power-law' case to the optimal `redundant code' case (the step function in Fig.~\ref{fig:spectrum}a), which helps protect the information from loss. To test this hypothesis we investigate the effect of higher and lower levels of dropout noise on the power spectrum (Fig.~\ref{fig:Noisy is non-linear}). 

In early layers, both the high and low dropout cases are well described by a simple power-law. In contrast, intermediate and late layers exhibit increasingly pronounced step-like spectra, especially in the higher-dropout layer. These properties indicate a large linear redundancy in the code, which is a characteristic signature of error-correcting codes.

Identifying the exact error-correcting code is beyond the scope of this paper, but we note that the non-linear decay is not a sharp step function, as would be the case with a linear error-correcting code. While this decay is consistent with a suboptimal encoding, we hypothesise that instead, the learnt code is not entirely linear. Consistently, the dropoff becomes more pronounced in later layers of the network, likely because linear error correction early in the network is both challenging (there are fewer layers to arrange the information into a code) and less necessary (as the image already contains many redundancies, minimising damage done by losing a couple of pixels).

\begin{figure}
    \centering
    \includegraphics{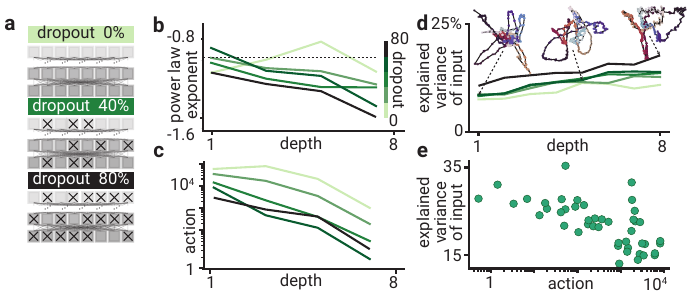}
    \caption{\textbf{Dropout during training leads to rapidly decaying eigenspectra and smooth random projections.} \textbf{a.} Schematic illustrating dropout across all layers with different strengths (0\% top, 40\% middle, 80\% bottom). \textbf{b.} Estimated power law exponent $\alpha$ as a function of network depth for different levels of dropout (shades of green). \textbf{c.} Average action of random projections as a function of network depth for different levels of dropout. \textbf{d.} Average percentage of input variance explained by random projections as a function of network depth for different layers of dropout. Inset shows three examples of random projections. \textbf{e.} Average percentage of input variance explained by random projections as a function of the average action of the random projection.}
    \label{fig:dropout 1}
\end{figure}

\textbf{Dropout incentivizes fast decay of eigenspectrum.} As a network's code depends on the amount of noise present (Fig.~\ref{fig:error}), it is natural to ask how the learned code changes as the dropout rate increases. We systematically varied the amount of dropout the network experiences (Fig.~\ref{fig:dropout 1}a) and observed that higher dropout generally produce lower power law exponents (Fig.~\ref{fig:dropout 1}b). Interestingly, already for a dropout rate of $20\%$, the power law exponent remains very close to $-1$, as in the visual cortex of the mouse\autocite{stringer2019high} and the macaque\autocite{kong2022increasing}. Consistent with our observation in networks without dropout (Fig.~\ref{fig:projection}), networks with lower power law exponent also tend to produce random projections with lower average action (Fig.~\ref{fig:dropout 1}c).

To test whether networks with higher dropout might encode more robust representations of their input throughout their layers, we computed the variance of the input explained by random projections from different layers of the network (see Methods). We found that the explained variance increases with both dropout rate and network depth (Fig.~\ref{fig:dropout 1}d). Consistently, we found that trajectories with low average action tend to explain a higher fraction of input variance (Fig.~\ref{fig:dropout 1}e).

These findings suggest that introducing dropout during training can reliably produce networks with a faster-decaying eigenspectrum. Random projections from these networks exhibit low average action and tend to resemble robustly and linearly encode some global features of the input.

\textbf{Efficient codes provide a framework for understanding language model interpretability.} The success of deep learning models in natural language processing, particularly the transformer architecture\autocite{vaswani2017attention}, provides a range of new challenges and opportunities for interpretability research\autocite{elhage2021mathematical,elhage2022toy,bills2023language}. A prominent challenge is the unprecedented size of large language models\autocite{brown2020language}, which makes the use of interpretability methods that rely on human annotation of individual components infeasible. A recent study proposes a GPT-assisted method for generating single neuron explanations in a scalable fashion\autocite{bills2023language}. While the proposed method manages to provide explanations with predictive power for thousands of neurons, the approach is limited to cases without polysemanticity and superposition, which our work suggests is a small fraction. To investigate whether our approach of interpreting random projections might be applicable to this setting, we turned to analyze the eigenspectrum of a large language model\autocite{radford2019language}.

\begin{figure}[b]
    \centering
    \includegraphics{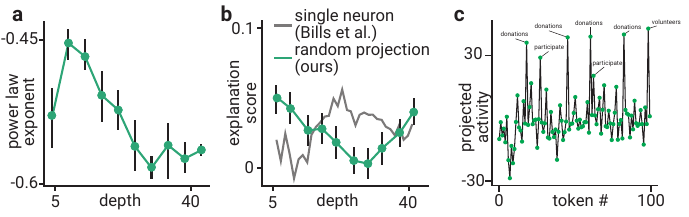}
    \caption{\textbf{Codes in large language models.} \textbf{a.} Estimated power law exponent $\alpha$ as a function of network depth of a large language model\autocite{radford2019language}. Error bars indicate 95\% confidence interval. \textbf{b.} Explanation score\autocite{bills2023language} as a function of network depth for explanations for the activation of single neurons (gray) or random projections (green). Data extracted from ref.~\cite{bills2023language}. Error bars indicate 95\% confidence interval. \textbf{c.} Illustrative example of a low-action random projection from layer 3. }
    \label{fig:dropout}
\end{figure}

We calculated the power law exponent $\alpha$ as a function of network depth. We observed that the power law exponent consistently remains above the critical value ($\alpha \geq -1$, Fig.~\ref{fig:dropout}a), indicating unbounded variance of random projections. This observation is consistent with the highly nonlinear nature of natural language\autocite{linzen-2016-issues}.

To investigate whether a lower power law exponent might nonetheless translate into more expressive random projections, we compute the explanation score (see Methods) for the activation of single neurons and random projections across different layers of the network (Fig.~\ref{fig:dropout}b,c). Interestingly, we found that the average explanation score changes non-monotonically, starting at comparatively high values in the first layers, decreasing up until two-thirds of the network depth, and eventually increasing again. The explanation score for random projections is the inverse of the score for single neurons, with our method performing well in the layers in which the single-neuron method performs poorly. We hypothesize that the random projection and single neuron approach may complement each other. By using both in tandem, we might obtain a more complete picture of coding in neural networks, bringing us close to the goal of understanding the outputs of large models.

\section{Discussion}

In this study, we explored the properties of neural network codes and their implications for interpretability, leveraging concepts from information theory and neuroscience. Our analysis revealed that the eigenspectrum of the activation covariance matrix provides insights into the redundancy of a network's code and its capacity for error-correction. Our results demonstrated that networks with higher dropout levels tend to employ codes with faster-decaying eigenspectra, yield smoother random projections, and better explanations of input variance. Furthermore, we found that our approach to interpreting random projections might complement existing single-neuron interpretability methods, particularly in the context of large language models. Overall, our findings provide a top-down perspective on neural network interpretability, advancing our understanding of the underlying structure of these models and suggesting new avenues for future research.

A fascinating aspect studying eigenspectrum decay in artificial neural networks is the potential connection to neuroscience\autocite{stringer2019high,kong2022increasing,wang2023scale}. In the same way that sufficiently large animals exhibit a neural code that optimally trades coding capacity and robustness to perturbations\autocite{stringer2019high,kong2022increasing}, we hypothesize that sufficiently large artificial neural networks might learn to strike the same balance. Indeed, a recent study indicates that networks with a fast eigenspectrum decay are more robust to adversarial examples and investigates which aspects of training and network architecture result in a more optimal decay\autocite{ghosh2022investigating}. A systematic study of how the eigenspectrum decay changes as a function of network size might reveal important information on whether we can expect large language models\autocite{brown2020language} to have smooth random projection.

An important direction for future research is to investigate the role of nonlinearities in the network code. Considering that neural networks were initially developed for nonlinear signal processing\autocite{lapedes1987nonlinear}, it is moderately surprising that modern networks allow linear read-out from their hidden layers\autocite{alain2016understanding,belrose2023eliciting}. Several factors incentivize linearity in hidden layers, such as the linear structure of the residual stream\autocite{he2016deep,elhage2021mathematical} and the training objective of supervised learning which incentivizes linear separability of concepts\autocite{bengio2013representation}. Conversely, linear codes can have comparable performance to nonlinear codes\autocite{mackay1997near}; hence they are not disadvantaged under optimization pressure. However, the exact reasons for which linear structure networks converge to\autocite{morcos2018insights} is an important open question into which our current work might provide insight.

Overall, our efficient channel coding framework provides a top-down perspective on interpretability in neural networks and offers a promising direction for future research in understanding deep learning models' underlying structure and behavior.

\section{Methods}

\textbf{Statistics.} We perform a Principal Component Analysis on the activations of the ReLu units of a given layer. To reduce the computational capacity required by very high-dimensional data, we apply a subsampling method by selecting every \texttt{s}-th component, where \texttt{s} is computed such that the resulting dimension is less than 8000. We employ a randomized SVD solver for PCA fitting and restrict the number of components to a maximum of 1000.

After fitting the PCA model, we perform a Huber regression analysis of the explained variance of the PCA components to estimate each layer's power-law exponent ($\alpha$). The Huber regression is used due to its robustness to outliers. Specifically, we fit the Huber regression model on the logarithm of the PCA components' indices (from 10 to 50) and the logarithm of their corresponding explained variances.

We use a recently published method for automated interpretability in language models\autocite{bills2023language}, which aims to explain the activation patterns of individual neurons in a large language model (GPT-2 XL) with a larger language model (GPT-4). We use GPT-4 to generate explanations of random projections of the activations of all neurons in a given layer of GPT-2. The value of these explanations is then assessed by a third model (GPT-3.5), which predicts the activation of the explained neuron on a held-out test set. The correlation between the expected and the real output is then the explanation score.

\textbf{Model training.} 
We studied the eigenspectrum of a ResNet-based Variational Autoencoder (ResNet-VAE) activations for image generation tasks. The ResNet-VAE model is built upon the ResNet-18 architecture pretrained on ImageNet. We remove the final fully connected layer from the ResNet architecture and replace it with two sets of linear layers to encode latent variables \texttt{mu} and \texttt{logvar}. The model employs two fully connected hidden layers with 1024 units each, followed by the latent embedding layer with 256 dimensions. A dropout rate of 0.2 is applied during training to promote regularization.

The model is trained on the CIFAR-10 dataset, which consists of 50,000 train and 10,000 test instances of 32x32 pixel images belonging to 10 different classes. The images are resized to 224x224 pixels and normalized before being fed into the network. We use the Adam optimizer with a learning rate of \texttt{1e-3} to optimize the network parameters. The training process consists of 20 epochs, with a batch size of 50. The model performance is measured by a custom loss function that combines Mean Squared Error (MSE) and Kullback-Leibler Divergence (KLD). 

In addition to the ResNet-VAE model, we studied the GPT-2 XL model \autocite{radford2019language} through the PyTorch Hub library. We computed explanation scores with the code published by Bills et al.~(ref.~\autocite{bills2023language}, but computed simulation scores with 
\texttt{text-davinci-003}
rather than GPT-4, since token logprobs are not available through the API at the time of this study.

\textbf{Code availability.} After publication code will be publically available as a GitHub repository; at this time, all experiments are available in the supplementary material.


{
\small
\printbibliography
}
\newpage

\end{document}


\maketitle

\section{Experimental details}
A zip file is provided in this supplementary material with the exact files used to create the results showcased in this experiment. We also provide brief descriptions of each file here to allow those looking to recreate our work or to find specific experiment details quicker access. After the anonymous peer review process is over these will be provided on a public github linked in the main text.

\textbf{ResNetVAE/modules.py}
This code defines a neural network model called ResNet\_VAE that is capable of encoding and decoding image data using a variational autoencoder architecture. The model uses a pre-trained ResNet18 model for encoding, and includes several fully connected layers and convolutional layers for decoding. The model is capable of generating a reconstructed image from an input image, as well as outputting a latent representation of the input image that can be used for further downstream analysis.

\textbf{ResNetVAE/ResNetVAE\_cifar10.py}
The ResNetVAE\_cifar10.py script contains the code to train a variant of the Variational AutoEncoder using the ResNet architecture as an encoder to encode an image into its latent representation. The script uses the CIFAR10 dataset to train the model and record the loss metrics throughout the training process. The code handles saving and loading of model and optimizer states, and uses weights and biases for logging and visualization. Finally, the trained models are stored in local disk space as well. The script has an adjustable dropout probability and can be trained for an adjustable number of epochs.

\textbf{ResNetVAE/make\_plots.py}
This code is used to analyze the properties of a ResNet-VAE neural network model trained on the CIFAR-10 dataset. The code first loads the model and uses hooks to extract activations at specified layers. The activations are then used to perform PCA analysis and the variance explained by each component is plotted. The code then generates activations for a set of dot images and project them onto a low-dimensional space obtained through a random projection matrix. The energy of the projection and the correlation between inputs and the projection are computed and plotted. Finally, the code computes a decay coefficient for each layer of the model and plots it against the energy of the projection and the correlation between inputs and the projection. The results are saved to disk as figures.

\textbf{ResNetVAE/helpers.py}
This code contains several utility functions for computer vision and machine learning tasks. It includes a function to generate a sequence of dot images moving in a circular path, and preprocesses the images using the default transform for CIFAR images. Another function computes the total energy of a set of points in n-dimensional space by calculating the squared magnitude of the differences between successive points. Additionally, there is a function that returns a dictionary of random projection matrices with specified dimensions for different layers in a neural network. These functions can be used to generate image data, calculate energy, and create random projection matrices for machine learning tasks.

\textbf{automated-interpretability/neuron-explainer/generate\_trajectory.py}
This code generates random trajectories for a pre-trained language model. It captures activations of certain layers during a forward pass and computes the total energy of a set of points in n-dimensional space. It then finds the minimum trajectory and project it to a 1D plot. The process is repeated multiple times to visualize different trajectories in each iteration. The final result is a set of PDF files and NPY files showing the 1D projections of different minimum trajectories. The generated plots can be used to gain insights into how the language model processes various sequences of text inputs.

\textbf{automated-interpretability/neuron-explainer/generate\_explanations.py}
This script generates explanations for the neuron activation patterns of a pre-trained GPT-2 XL language model using a calibrated simulator. The explanations are generated for the top 10 activation patterns of each layer in the neural network based on their mean maximum activation value compared to all activation patterns. The TokenActivationPairExplainer is used to generate explanations for the neuron activations, followed by simulating and scoring the explanation using an UncalibratedNeuronSimulator. The generated explanations and scores are saved for later use.

\begin{figure}
    \centering
    \includegraphics{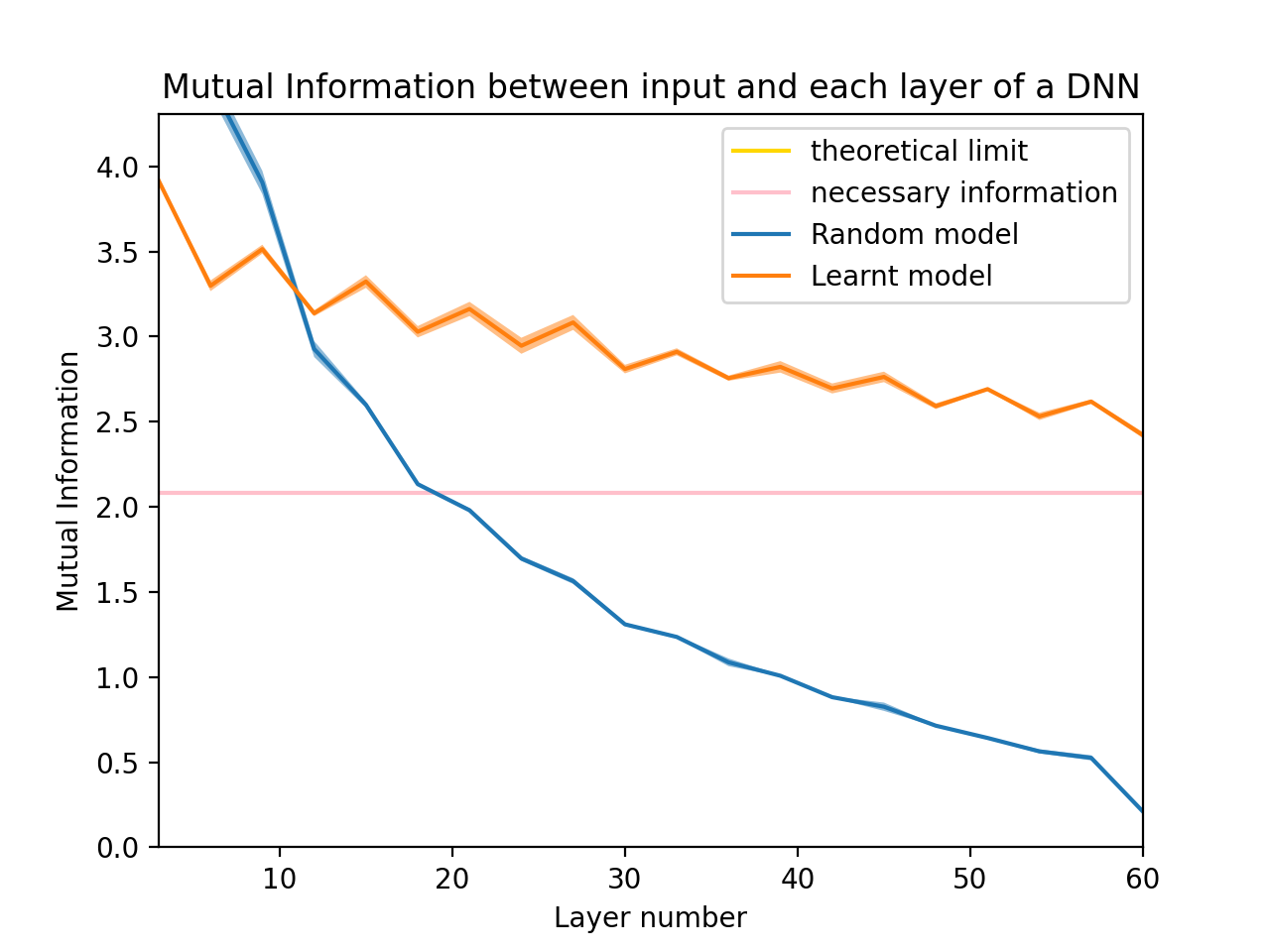}
    \caption{MINE is deployed to estimate the continuous mutual information between the input and a given layer of a neural network. One neural network is trained in the presence of dropout to preserve a subset of the information in the input while the other isn't. This demonstrates that in some settings a neural network is capable of learning an error correcting scheme, the random model provides a demonstration of what happens to information in this scenario that isn't intentionally preserved}
    \label{fig:error correction}
\end{figure}

\textbf{Estimate-MI-of-NN-Layers.py}
This script trains a very deep neural network on a simple task (preserving information) in a regime where error correction is needed to preserve any information. The results of this program are shown in figure \ref{fig:error correction}

\section{Theorems and Proofs}
We provide formal statements and proofs behind the results claimed in the main text.

\subsection{Optimal sparse input ratio}
This subsection derives the predicted optimal ratio \# dimensions-per-feature to sparsity.

First, we formally define the sparse input distribution. It is based on a uniform random distribution with $S$ features set to 0.

\begin{definition}[Sparsification]
Let $S$-sparsification be a process that takes a vector random variable $v$ and returns a random variable $v'$ with each element of $v'$ independently set to 0 with probability $S$.
If $X$ is any distribution with $i.i.d.$ elements then the $S$-sparsification of $X$ is the distribution of $X$ after $S$-sparsification, denote it as $X^S$
\end{definition}

The entropy of such a distribution is

\begin{lemma}[Entropy of sparse distribution]
Let $X$ be some distribution. If $X^S$ is the distribution after $S$-sparsification then the entropy of $X^S$ is:
$$
H(X^S) = -n S \log(S) - n(1-S)\log(1-S) + n(1-S)H_{elem}(X)
$$ 
Where $H_{elem}(X)$ is the elementwise entropy of $X$
\end{lemma}
\begin{proof}
Express the joint entropy as the sum of the individual entropies (chain rule of entropy)
$$
H(X^S)=H(X^S_0, X^S_1, \ldots, X^S_n) = \sum_i (H^S_i|X^S_0,\ldots, X^S_{i-1}, X^S_{i+1}, \ldots X^S)
$$
as elements are i.i.d. (a condition of sparsification).
$$
H(X^S) = \sum_i (H^S_i) = n H(X^S_0)
$$
W.L.O.G. we have set all entropies equal to that of the first element of the vector. 

If we assume smoothness of $X$ the probability that any un-sparcified element of a vector is exactly 0 is vanishingly small $P(X_0=0)\approx 0$, allowing us to assume that if $X_0^S$ is 0, it has been set this way by sparsification. This implies that a single element of the sparcified vector has the distribution:
$$
P(X^S_0=Y)=
\begin{cases}
    (1-S)P(X_0^S = Y), & \text{if } Y \neq 0 \\
    S, & \text{if } Y=0
\end{cases}
$$
Directly calculating the entropy of this distribution:
$$
 H(X^S_0) = - \sum_{X^S_0} P(X^S_0) \log( P(X^S_0)) = - \sum_{X^S_0\neq 0} P(X^S_0) \log( P(X^S_0)) + P(X^S_0=0) \log( P(X^S_0=0))
$$
$$
 H(X^S_0) \approx -\sum_{X^S_0\neq 0} (1-S) P(X_0) \log( (1-S)P(X_0)) - S \log(S)
$$
$$
 H(X^S_0) \approx - (1-S)\sum_{X^S_0\neq 0} P(X_0) \log( P(X_0)) - (1-S)\sum_{X^S_0\neq 0} P(X_0) \log( 1-S) - S \log(S)
$$
$$
 H(X^S_0) \approx - (1-S)H(X_0) - (1-S)\log( 1-S) - S \log(S)
$$
Which proves the result of the lemma.
\end{proof}

The above statement is approximate, taking the assumption that $P(X_0=0)$ is arbitrarily small. This can be turned into a strongly rigorous statement by taking smoothness or other assumptions which amount to bounding $P(X_0=0)<\varepsilon$. In this case the stated entropy result is within additive $\epsilon$. For the case we are interested in (Elhage et al.) $X$ is a uniform distribution of high precision floats, making $P(X_0=0)\approx0$ sufficiently accurate.

The stated ratio simply follows from this lemma by assuming the precision (and hence entropy) of the internal state/hidden layer/linear algebra is equal to that of the unsparcified input distribution, $X$: $H_elem$. The channel capacity of $m$ hidden units is $m\times H_elem$. The ratio, $m:n$, is found by divding one by the other.

\subsection{Hardness of decoding a certain problem}
As codes become more efficient (approach the Shannon limit given the entropy) it seems likely that they become less interpretable. To provide weight to this conjecture we outline a learning problem such that any attempt at decoding \footnote{i.e. to take the code of an input and resolve the original input, or what information about the input has been conserved} the optimal code in the final layer in polynomial time must fail, assuming cryptographic primitives. 

The existence of such a task is relatively trivial, we can see that a neural network is capable of learning to implement a discrete exponential, an function which is easy one-way but complicated to find the inverse of. If any algorithm were able to decode the input from a given output we would be able to use the same algorithm to find the inverse, i.e. to solve the discrete logarithm problem. The optimality of code constraint forces the model to keep only the bits it needs for the problem, leaving just the exponential and no ``leaks" as to what the input was.

Formally the problem gives 3 values as input: a prime, $p$, a primitive element, $g$ of $\mathbb{Z}^*_p=\{1, \ldots, p-1\}$, and $x\in\mathbb{Z}^*_p$. The tasks is then to output $g^x (\text{mod} p)$. This is an easy function to implement and very likely learnable up to large $p$, $g$, and $x$ (only existence, not learnability, is strictly needed to show the result). It is now obvious how any algorithm that could take the output of this circuit and calculate $x$ could solve DLP.

DLP is thought to be average-case hard for chosen groups (chosen $p$), restricting our problem case to these instances completes our result. 
Many other cryptographic one-way functions could be used in the place of DLP here, strengthening this result to rely on any one of a number of possible assumed hardness results. 
Indeed, given a public-private key encryption scheme which is average case hard for every bit of the input, we can modify the learning problem such that the neural network implements the public key encryption, again any decoder that breaks even a single bit of the encrypted activations in polynomial time could be used to recover the original message without access to the private key.
\subsection{Channel capacity of a dropout layer}

This subsection proves the channel capacity result claimed in the main text, namely that a neural network with $n$ dropout layers, with dropout rate $D$ has a channel capacity of $C_{\mathit{dropout}}\sim C (1-D)$, where $C$ is the capacity of the dropout-less layer. 

\begin{theorem}
The channel capacity of $m$ independent neurons, each with capacity $C_\text{base}$ when subject to an uncorrelated dropout layer with rate $D$ is bounded by:

$$
 (1-\varepsilon)\times(1-D)\times C_\text{base} \times m\leq
 C^m_\text{dropout}
 \leq (1-D)\times C_\text{base} \times m,
$$

where $\varepsilon = -\log(1-2^{-C_\text{base}})$ tends to zero asymptotically for large $C_\text{base}$ (high precision).
\end{theorem}
\begin{proof}
We first note that $C^{m}_\text{dropout}$ is equivalent to $m\times C^{1}_\text{dropout}$ due to the chain rule of entropy and the independence of channels. 

We then establish the lower and upper bounds separately for $C^{1}_\text{dropout}$, both rely on the similarity of this channel to the standard erasure channel. 

\textbf{Lower bound}(proof by construction) Create a uniform distribution on all but the 0 symbol, call this $X$. The entropy of this distribution is equal to the full uniform distribution sans 1 state:
$$
H(X) = \log(2^{C^1_{base}}-1) = \log(1-2^{-C^1_{base}})+C^1_{base}
$$

Let $Y$ be the distribution given by putting $X$ through the dropout layer. The mutual information between the two is:

$$
I(X;Y) = H(X)-H(X|Y) = H(X) - \sum P(Y=y) H(X|Y=y)
$$

If $y=0$, $ H(X|Y=y) = H(X)$ as dropout applies to all inputs equally and the 0 symbol in $X$ has no probability mass, if $y\neq0$ then $y=x$ and $ H(X|Y=y)=0$. Thus:

$$
I(X;Y) = H(X)-H(X|Y) = H(X) - P(Y=0) H(X)=(1-D)H(X)
$$

Implying the lower bound on the channel capacity

$$
C^{1}_\text{dropout} \geq (1-D)\times(\log(1-2^{-C^1_{base}})+C^1_{base})
$$

\textbf{Upper bound}(proof by contradiction) Assume toward contradiction that $C^{1}_\text{dropout} = (1-D)C_{base}^1+\varepsilon$ for some fixed $\varepsilon>0$. Let the n-bit erasure channel be the extension of the binary erasure channel to an input of $n$ bits, i.e. on input $x \in \mathcal{X}=\{ 0,1\}^n$ the channel produces output $y \in \mathcal{Y}=\{ 0,1\}^n+{e}$, where $e$ is an error symbol, with probability:
$$
P(Y=y|X=x) = 
\begin{cases}
    \alpha, & y = e \\
    1-\alpha, & y=x
\end{cases}
$$
The proof of the channel capacity of this channel is a trivial extension of the binary case:
$I(X;Y)=H(X)+H(X|Y)= H(X) - \sum P(Y=y) H(X|Y=y) = (1-\alpha) H(X)$

Which is maximized when $X$ is uniform, giving a capacity of $(1-\alpha)n = (1-\alpha)C_{base}$.

However, given access to a $n$-bit erasure channel with $\alpha =D$ we can recreate a dropout channel by setting all $e$ symbols to 0. If The channel capacity of our dropout layer was as assumed we could use the encoding procedure for the dropout layer, pass it through the $n$-bit erasure channel, perform the $e=0$ operation and achieve the dropout layer capacity. As $\varepsilon>0$ this implies it is possible to transmit more information than the channel capacity through the erasure channel, causing a contradiction.
\end{proof}

\subsection{Information loss from power law distribution}
The entropy of a distribution with variance following power law distribution is highly limited, constraining the channel capacity ($C=H(X)-H(X|Y)<H(X)$). This section calculates how much information is not transmitted by the power law distribution compared to a uniform (flat distribution).

\begin{theorem}
    The entropy of a distribution of $m$ neurons to precision $n$ following a power law distribution with exponent $\alpha$, $X$, is upper bounded by:
    $$
    H(X) \lessapprox m n - \alpha m \log(m/e)
    $$
    if the distributions of each number follow a normal distribution.
\end{theorem}
\begin{proof}
    W.L.O.G. Assume the variances, $\sigma_i$, are normalised such that $\sigma_1 = 1$. The power law distribution then implies $\sigma_i = i^\alpha$
    
    Entropy is upper bounded by the case where all $m$ channels are independent:
    $$
    H(X)\leq \sum_i^m H(X_i)
    $$

    Assuming the first element optimally uses every bit it would uniformly distribute over all values, the subsequent variances would take up a smaller and smaller section (as the variance must decrease, the entropy implied by each variance is approximately:
    \begin{equation} \label{eq: unidistro}
        H(X_i) \approx \log(2^n \sigma_i)
    \end{equation}
    
    Substituting in variance gives:
$$
    H(X)\leq \sum_i^m \log(2^n \sigma_i) = \sum_i^m \log(2^n i^\alpha) = m n - \alpha \log(\prod^m_i i)
$$  
    Applying Stirlings lemma gives the final result:
    $$
    H(X)\lessapprox m n - \alpha m \log(m/e)
    $$
\end{proof}

Considering the Gaussian distribution has the highest entropy for a given variance the choice to assume a normal distribution in the previous answer may appear strange. Normal was chosen so that the first variance could be assumed to be an optimally dense code, if Gaussian is chosen edge effects (i.e. running out of numbers for a true Gaussian) must be considered, making it unclear what the initial value of $\sigma_0$ is. However, a reader that wishes to use a Gaussian may simply modify the above proof to replace the entropy in the equation \ref{eq: unidistro} with that of a Gaussian. Any reasonable values of the variance result in an additive term of at most $\frac{m}{2}log(2 \pi e)$, making the normal approximation close to the Gaussian.

\subsection{Connection between power law exponent and energy of a random projection}
This section connects the action with the power series decay parameters, $\alpha$.,

We're given a vector of hidden activations, $A \in \mathbb{R}^m$, centered around zero, $A \sim \mathcal{N}(0,\Sigma)$, and random projections, $r\in\mathbb{R}^{m\times 3}$ centered at 0 with normalised variance $r_i\sim\mathcal{N}(0,\frac{1}{m}I)$. Expand definition of the action:
$$
    \langle E[z] \rangle = \frac{1}{2} \langle ||A r||^2 \rangle = \frac{1}{2} \langle r^TA^TAr \rangle.
$$
The action is a scalar, thus it is equal to its trace, applying the cyclic and linear properties of the trace we rearrange and pull the expected value into the trace (the trace trick):
$$
    \langle E[z] \rangle = \frac{1}{2} \langle \text{tr}(r^Tr) \rangle  \text{tr}(\langle A^TA \rangle) = \frac{1}{2} \text{tr}(\Sigma) 
$$

As the trace is equal to the sum of the eigenvalues, and the eigenspectrum of $\Sigma$ is the set of variances, we rewrite the expected action as a sum over the power law,
$$
    \langle E[z] \rangle =  \frac{1}{2} \sum_i \lambda_i = \frac{1}{2} \sum_{i} i^{\alpha} \approx \frac{1}{2} \sum_{i=1}^\infty i^{\alpha} = \frac{1}{2} \sum_{i=1}^\infty \frac{1}{i^{-\alpha}} = \frac{1}{2} \zeta({-\alpha}).
$$
As the later terms of the sum rapidly approach zero, we can approximate the sum with the infinite series. Rewriting the expression in terms of $-\alpha$ then results in the connection with the $\zeta$ function,
$$
    \langle E[z] \rangle \approx \frac{1}{2} \sum_{i=1}^\infty i^{\alpha} = \frac{1}{2} \sum_{i=1}^\infty \frac{1}{i^{-\alpha}} = \frac{1}{2} \zeta({-\alpha}).
$$